\documentclass{article}
\PassOptionsToPackage{table}{xcolor}
\usepackage[preprint]{log_2022}  %

\usepackage{booktabs}            %
\usepackage{multirow}            %
\usepackage{amsfonts}            %
\usepackage{graphicx}            %
\usepackage{duckuments}          %
\usepackage{subcaption}

\usepackage[hidelinks,backref=page]{hyperref}
\usepackage{amsbsy,amsmath,amsfonts,amsthm,amssymb, mathdots,mathtools, latexsym, bm}

\usepackage[nameinlink, noabbrev, capitalize]{cleveref}
\usepackage[numbers,compress,sort]{natbib}

\usepackage{tikz}           %
\usetikzlibrary{arrows.meta} %
\usepackage{tikz-feynman}   %
\usepackage{todonotes}      %
\usepackage{comment}
\usepackage{enumitem}

\definecolor{rwth-blue}{cmyk}{1,.5,0,0}\colorlet{rwth-lblue}{rwth-blue!50}\colorlet{rwth-llblue}{rwth-blue!25}
\definecolor{rwth-violet}{cmyk}{.6,.6,0,0}\colorlet{rwth-lviolet}{rwth-violet!50}\colorlet{rwth-llviolet}{rwth-violet!25}
\definecolor{rwth-purple}{cmyk}{.7,1,.35,.15}\colorlet{rwth-lpurple}{rwth-purple!50}\colorlet{rwth-llpurple}{rwth-purple!25}
\definecolor{rwth-carmine}{cmyk}{.25,1,.7,.2}\colorlet{rwth-lcarmine}{rwth-carmine!50}\colorlet{rwth-llcarmine}{rwth-carmine!25}
\definecolor{rwth-red}{cmyk}{.15,1,1,0}\colorlet{rwth-lred}{rwth-red!50}\colorlet{rwth-llred}{rwth-red!25}
\definecolor{rwth-magenta}{cmyk}{0,1,.25,0}\colorlet{rwth-lmagenta}{rwth-magenta!50}\colorlet{rwth-llmagenta}{rwth-magenta!25}
\definecolor{rwth-orange}{cmyk}{0,.4,1,0}\colorlet{rwth-lorange}{rwth-orange!50}\colorlet{rwth-llorange}{rwth-orange!25}
\definecolor{rwth-yellow}{cmyk}{0,0,1,0}\colorlet{rwth-lyellow}{rwth-yellow!50}\colorlet{rwth-llyellow}{rwth-yellow!25}
\definecolor{rwth-grass}{cmyk}{.35,0,1,0}\colorlet{rwth-lgrass}{rwth-grass!50}\colorlet{rwth-llgrass}{rwth-grass!25}
\definecolor{rwth-green}{cmyk}{.7,0,1,0}\colorlet{rwth-lgreen}{rwth-green!50}\colorlet{rwth-llgreen}{rwth-green!25}
\definecolor{rwth-cyan}{cmyk}{1,0,.4,0}\colorlet{rwth-lcyan}{rwth-cyan!50}\colorlet{rwth-llcyan}{rwth-cyan!25}
\definecolor{rwth-teal}{cmyk}{1,.3,.5,.3}\colorlet{rwth-lteal}{rwth-teal!50}\colorlet{rwth-llteal}{rwth-teal!25}
\definecolor{rwth-gold}{cmyk}{.35,.46,.7,.35}
\definecolor{rwth-silver}{cmyk}{.39,.31,.32,.14}

\usepackage{wrapfig}

\theoremstyle{definition}

\renewcommand{\phi}{\varphi}
\renewcommand{\epsilon}{\varepsilon}

\definecolor{rwth-blue}{cmyk}{1,.5,0,0}\colorlet{rwth-lblue}{rwth-blue!50}\colorlet{rwth-llblue}{rwth-blue!25}
\definecolor{rwth-violet}{cmyk}{.6,.6,0,0}\colorlet{rwth-lviolet}{rwth-violet!50}\colorlet{rwth-llviolet}{rwth-violet!25}
\definecolor{rwth-purple}{cmyk}{.7,1,.35,.15}\colorlet{rwth-lpurple}{rwth-purple!50}\colorlet{rwth-llpurple}{rwth-purple!25}
\definecolor{rwth-carmine}{cmyk}{.25,1,.7,.2}\colorlet{rwth-lcarmine}{rwth-carmine!50}\colorlet{rwth-llcarmine}{rwth-carmine!25}
\definecolor{rwth-red}{cmyk}{.15,1,1,0}\colorlet{rwth-lred}{rwth-red!50}\colorlet{rwth-llred}{rwth-red!25}
\definecolor{rwth-magenta}{cmyk}{0,1,.25,0}\colorlet{rwth-lmagenta}{rwth-magenta!50}\colorlet{rwth-llmagenta}{rwth-magenta!25}
\definecolor{rwth-orange}{cmyk}{0,.4,1,0}\colorlet{rwth-lorange}{rwth-orange!50}\colorlet{rwth-llorange}{rwth-orange!25}
\definecolor{rwth-yellow}{cmyk}{0,0,1,0}\colorlet{rwth-lyellow}{rwth-yellow!50}\colorlet{rwth-llyellow}{rwth-yellow!25}
\definecolor{rwth-grass}{cmyk}{.35,0,1,0}\colorlet{rwth-lgrass}{rwth-grass!50}\colorlet{rwth-llgrass}{rwth-grass!25}
\definecolor{rwth-green}{cmyk}{.7,0,1,0}\colorlet{rwth-lgreen}{rwth-green!50}\colorlet{rwth-llgreen}{rwth-green!25}
\definecolor{rwth-cyan}{cmyk}{1,0,.4,0}\colorlet{rwth-lcyan}{rwth-cyan!50}\colorlet{rwth-llcyan}{rwth-cyan!25}
\definecolor{rwth-teal}{cmyk}{1,.3,.5,.3}\colorlet{rwth-lteal}{rwth-teal!50}\colorlet{rwth-llteal}{rwth-teal!25}
\definecolor{rwth-gold}{cmyk}{.35,.46,.7,.35}
\definecolor{rwth-silver}{cmyk}{.39,.31,.32,.14}

\title[Reassessing the Long-Range Graph Benchmark]{Where Did the Gap Go?\\ Reassessing the Long-Range Graph Benchmark}

\author[Tönshoff et al.]{%
Jan Tönshoff \thanks{corresponding author}\\
\institute{RWTH Aachen}\\
\email{toenshoff@informatik.rwth-aachen.de}\And
Martin Ritzert\\
\institute{Georg-August-Universität Göttingen}\\
\email{ritzert@informatik.uni-goettingen.de} \And
Eran Rosenbluth\\
\institute{RWTH Aachen}\\
\email{rosenbluth@informatik.rwth-aachen.de} \And
Martin Grohe\\
\institute{RWTH Aachen}\\
\email{grohe@informatik.rwth-aachen.de}
}

\begin{document}

\maketitle

\begin{abstract}
The recent Long-Range Graph Benchmark (LRGB, \hyperlink{cite.dwivedi2022long}{Dwivedi et al. 2022}) introduced a set of graph learning tasks strongly dependent on long-range interaction between vertices.
Empirical evidence suggests that on these tasks Graph Transformers significantly outperform Message Passing GNNs (MPGNNs).
In this paper, we carefully reevaluate multiple MPGNN baselines as well as the Graph Transformer GPS (\hyperlink{cite.rampavsek2022recipe}{Rampášek et al. 2022}) on LRGB.
Through a rigorous empirical analysis, we demonstrate that the reported performance gap is overestimated due to suboptimal hyperparameter choices.
It is noteworthy that across multiple datasets the performance gap completely vanishes after basic hyperparameter optimization.
In addition, we discuss the impact of lacking feature normalization for LRGB's vision datasets and highlight a spurious implementation of LRGB's link prediction metric. %
The principal aim of our paper is to establish a higher standard of empirical rigor within the graph machine learning community.
\end{abstract}

\section{Introduction}
Graph Transformers (GTs) have recently emerged as popular alternative to conventional Message Passing Graph Neural Networks (MPGNNs) which dominated deep learning on graphs for years.
A central premise underlying GTs is their ability to model long-range interactions between vertices through a global attention mechanism.
This could give GTs an advantage on tasks where MPGNNs may be limited through phenomenons like over-smoothing, over-squashing, and under-reaching, thereby justifying the significant runtime overhead of self-attention.

The Long-Range Graph Benchmark (LRGB) has been introduced by \citet{dwivedi2022long} as a collection of five datasets with strong dependence on long-range interactions between vertices:

\begin{itemize}[topsep=0pt, partopsep=0pt, parsep=0pt, itemsep=2pt]
    \item \emph{Peptides-func} and \emph{Peptides-struct} are graph-level classification and regression tasks, respectively.
    Their aim is to predict various properties of peptides which are modelled as molecular graphs.
    \item \emph{PascalVOC-SP} and \emph{COCO-SP} model semantic image segmentation as a node-classification task on superpixel graphs. %
    \item \emph{PCQM-Contact} is a link prediction task on molecular graphs. The task is to predict pairs of vertices which are distant in the graph but in contact in 3D space. 
\end{itemize}

The experiments provided by \citet{dwivedi2022long} report a strong performance advantage of GTs over the MPGNN architectures GCN \cite{kipf2017semi}, GINE \cite{hu2020pretraining}, and GatedGCN \cite{bresson2017residual}, in accordance with the expectations.
Subsequently, GPS
\cite{rampavsek2022recipe} reached similar conclusions on LRGB.
We note that these two works are strongly related and built on a shared code base.
Newer research on GTs (see \cref{rw}) is commonly based on forks of this code base and often cites the baseline performance reported by \citet{dwivedi2022long} to represent MPGNNs.

Our contribution is three-fold\footnote{The source code is provided here: \url{https://github.com/toenshoff/LRGB}}: 
First, we show that the three MPGNN baselines GCN, GINE, and GatedGCN all profit massively from further hyperparameter tuning, reducing and even closing the gap to graph transformers on multiple datasets. 
In fact, GCN yields state-of-the-art results on Peptides-Struct, surpassing several newer graph transformers.
On this dataset in particular, most of the performance boost is due to a multi-layer prediction head instead of a linear one, again highlighting the importance of hyperparameters.
Second, we show that on the vision datasets PascalVOC-SP and COCO-SP normalization of the input features is highly beneficial.
We argue that, as in the vision domain, feature normalization should be the default setting.
Third and last we take a closer look at the MRR metric used to evaluate PCQM-Contact.
There, we demonstrate different filtering strategies have a major impact on the results and must be implemented exactly to specification to facilitate reliable comparisons.

\subsection{Related Work}
\label{rw}
Our primary focus are the commonly used MPGNNs GCN \cite{kipf2017semi}, GINE \cite{hu2020pretraining}, and GatedGCN \cite{bresson2017residual} as well as the graph transformer GPS \cite{rampavsek2022recipe}.
There are many more MPGNN architectures \cite{DBLP:journals/corr/HamiltonYL17,xu2018powerful,chen2020simple,corso2020principal}, as well as graph transformers \cite{dwivedi2020generalization, ying2021transformers, kreuzer2021rethinking, shi2020masked, park2022grpe, weis2021self, rampavsek2022recipe, shirzad2023exphormer, kim2022pure, ma2023GraphInductiveBiases, pmlr-v202-he23a}, see also the survey by \citet{min2022transformer}.
Many newer graph transformer architectures have reported results on LRGB datasets, including Exphormer \cite{shirzad2023exphormer}, GRIT \cite{ma2023GraphInductiveBiases} and Graph ViT / GraphMLPMixer \cite{pmlr-v202-he23a}.
Several other graph learning approaches not based on transformers have also conducted experiments on LRGB, including  CRaWl \cite{crawl} and DRew \cite{pmlr-v202-gutteridge23a}.
Finally, we do see a connection of our work to graph learning benchmarking projects \cite{dwivedi2020benchmarkgnns,hu2020ogb} that also advocate for rigorous testing of graph learning architectures.

\section{Concerns}
\label{sec:concerns}

\begin{figure}[t]
    \begin{subtable}{0.48\textwidth}
        
\setlength{\tabcolsep}{3pt}
\begin{small}
\resizebox{\textwidth}{!}{
\textsc{
\begin{tabular}{l@{\hspace{4pt}}l@{\hspace{2pt}}c@{\hspace{4pt}}l@{\hspace{7pt}}c@{\hspace{4pt}}l@{\hspace{3pt}}}
\toprule
&Method     & \multicolumn{2}{c}{PEPTIDES-FUNC} & \multicolumn{2}{c}{PEPTIDES-STRUCT} \\
&           & Test AP $\uparrow$ & \!\!\!\!rel imp\!\!\!\!\!\!& Test MAE $\downarrow$ & \rlap{\!\!rel imp\!\!\!}\\
\midrule
\multirow{6}*{\rotatebox{90}{LRGB}}
&\hyperlink{cite.kipf2017semi}{GCN}                      & 0.5930 ± 0.0023 & & 0.3496 ± 0.0013 & \\
&\hyperlink{cite.Keyulu18}{GINE}                         & 0.5498 ± 0.0079 & & 0.3547 ± 0.0045 & \\
&\hyperlink{cite.bresson2017residual}{GatedGCN}          & 0.6069 ± 0.0035 & & 0.3357 ± 0.0006 & \\
&\hyperlink{cite.dwivedi2020generalization}{Transformer} & 0.6326 ± 0.0126 & & 0.2529 ± 0.0016 & \\
&\hyperlink{cite.kreuzer2021rethinking}{SAN}             & 0.6439 ± 0.0075 & & 0.2545 ± 0.0012 & \\
&\hyperlink{cite.rampavsek2022recipe}{GPS}               & 0.6535 ± 0.0041 & & 0.2500 ± 0.0005 & \\
\midrule
\multirow{4}*{\rotatebox{90}{Ours}}
&GCN                                                     & 0.6860 ± 0.0050 &\cellcolor{rwth-green!48}+16\% & \textbf{0.2460 ± 0.0007} &\cellcolor{rwth-green!90}+30\%               \\
&GINE                                                    & 0.6621 ± 0.0067 &\cellcolor{rwth-green!60}+20\% & 0.2473 ± 0.0017 &\cellcolor{rwth-green!90}+30\%                        \\
&GatedGCN                                                & 0.6765 ± 0.0047 &\cellcolor{rwth-green!33}+11\% & 0.2477 ± 0.0009 &\cellcolor{rwth-green!78}+26\%                        \\
&GPS                                                     & 0.6534 ± 0.0091 &\cellcolor{rwth-green!0}±\phantom{0}0\% & 0.2509 ± 0.0014 &\cellcolor{rwth-green!0}±\phantom{0}0\%      \\
\midrule
\multirow{6}*{\rotatebox{90}{Others}}
&\hyperlink{cite.crawl}{\textsc{CRaWl}}                                          & 0.7074 ± 0.0032 & & 0.2506 ± 0.0022 & \\
&\hyperlink{cite.pmlr-v202-gutteridge23a}{DRew}               & \textbf{0.7150 ± 0.0044} & & 0.2536 ± 0.0015 & \\
&\hyperlink{cite.shirzad2023exphormer}{Exphormer}             & 0.6527 ± 0.0043 & & 0.2481 ± 0.0007 & \\
&\hyperlink{cite.ma2023GraphInductiveBiases}{GRIT}            & 0.6988 ± 0.0082 & & \textbf{0.2460 ± 0.0012} & \\
&\hyperlink{cite.cite.pmlr-v202-he23a}{Graph ViT}             & 0.6942 ± 0.0075 & & \textbf{0.2449 ± 0.0016} & \\
&\hyperlink{cite.pmlr-v202-he23a}{G-MLPMixer}         & 0.6921 ± 0.0054 & & 0.2475 ± 0.0015 & \\
\bottomrule
\end{tabular}
}
}
\end{small}

        \caption{{Hyperparameter tuning on the peptides datasets. Best results (within stdev) in \textbf{bold}.}}
        \label{peptides_results}
    \end{subtable}
    \hfill
    \begin{subfigure}{0.51\textwidth}
        \includegraphics[width=\textwidth]{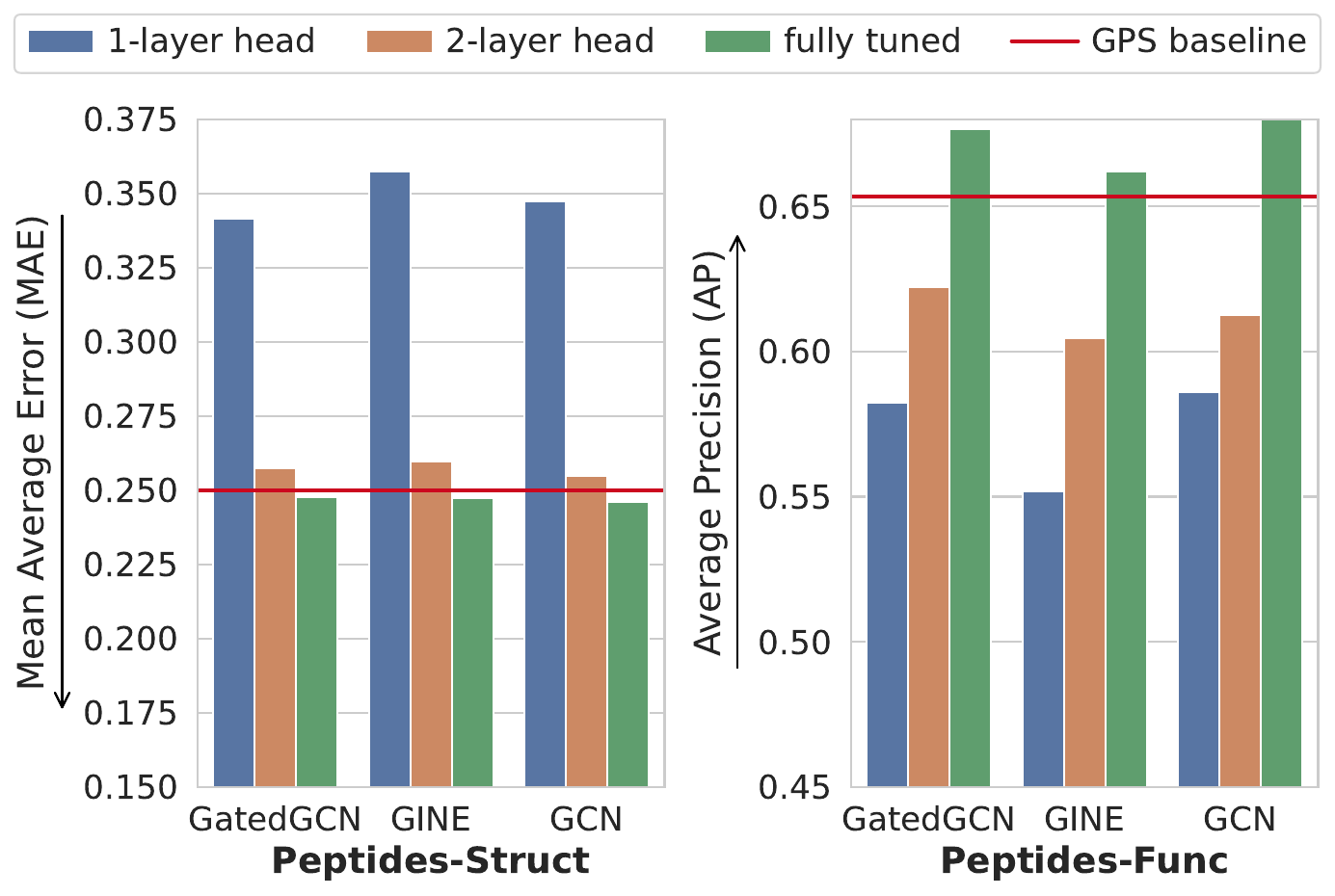}
        \caption{Exchanging the linear prediction head by an MLP accounts for most of the additional performance of all three MPGNNs, especially on Peptides-Struct}    
        \label{fig:peptides_head_ablation}
    \end{subfigure}
    \caption{On both Peptides datasets, all three MPGNNs surpass GPS. On Peptides-Struct a basic GCN model even achieves SOTA results.}
\end{figure}

\paragraph{Hyperparameters}
In this paper, we argue that the results reported by \citet{dwivedi2022long} are not representative for MPGNNs and suffer from suboptimal hyperparameters.
We provide new results for the same MPGNN architectures that are obtained after a basic hyperparameter sweep.
We tune the main hyperparameters (such as depth, dropout rate, \dots) in pre-defined ranges while strictly adhering to the official 500k parameter budget.
The exact hyperparameter ranges and all final configurations are provided in \cref{appendix:hp}.
As a point of reference, we reevalute GPS in an identical manner and also achieve significantly improved results on three datasets with this Graph Transformer.
The results reported for GPS may therefore also be subject to suboptimal configurations.
Note that we also view the usage of positional or structural encoding (none / LapPE \cite{dwivedi2020generalization} / RWSE \cite{dwivedi2021graph}) as a hyperparameter that is tuned for each method, including all MPGNNs.

\paragraph{Feature Normalization}
The vision datasets PascalVOC-SP and COCO-SP have multi-dimensional node and edge features with values spanning different orders of magnitude for different feature channels.
Passing this input to a neural network without channel-wise normalization can cause poorly conditioned activations.
While feature normalization is standard practice in deep learning and computer vision in particular, neither \citet{dwivedi2022long} nor any subsequent works using LRGB utilize it, except \textsc{CRaWl} \cite{crawl}.
We apply channel-wise linear normalization to all input features and show that all models (baselines and GPS) profit from it in an ablation in \cref{Fig:PascalNorm}.

\paragraph{Link Prediction Metrics}
The evaluation metric on the link-prediction dataset PCQM-Contact \cite{dwivedi2022long} is the Mean Reciprocal Rank (MRR) in a \emph{filtered} setting, as defined by \citet{NIPS2013_1cecc7a7}.
For predicted edge scores the MRR measures how a given true edge $(h,t)$ is ranked compared to all possible candidate edges $(h,x)$ of the same head.
As there might be multiple true tails $t$ for each head $h$, the \emph{filtered} MRR removes those other true tails (false negatives) from the list of candidates before computing the metric.
This filtering avoids erroneously low MRR values due to the model preferring other true edges and is common in link-prediction tasks.
Even though \citet{dwivedi2022long} explicitly define the metric to be the filtered MRR, the provided code computes the \emph{raw} MRR, i.e. keeping other true tails in the list. %
We report results on PCQM-Contact in a corrected filtered setting.
We additionally provide results with an extended filtering procedure where self-loops of the form $(h,h)$ are also removed from the set of candidates, since these are semantically meaningless and never positive. %
This is impactful as the scoring function used by \citet{dwivedi2022long} is based on a symmetric dot-product and therefore exhibits a strong bias towards self-loops.

\section{Experiments}
\label{sec:results}

\begin{figure}[t]
    \centering
    \begin{subtable}{0.47\textwidth}
        \setlength{\tabcolsep}{3pt}

\begin{small}
\resizebox{\textwidth}{!}{
\textsc{
\begin{tabular}{l@{\hspace{4pt}}l@{\hspace{2pt}}c@{\hspace{4pt}}l@{\hspace{7pt}}c@{\hspace{4pt}}l@{\hspace{3pt}}}
\toprule
&Method     & \multicolumn{2}{c}{PASCALVOC-SP} & \multicolumn{2}{c}{COCO-SP} \\
&           & Test F1 $\uparrow$ & \!\!\!\!rel imp\!\!\!\!\!\!& Test F1 $\uparrow$ & \rlap{\!\!rel imp\!\!\!}\\
\midrule
\multirow{6}*{\rotatebox{90}{LRGB}}&\hyperlink{cite.kipf2017semi}{GCN}                       & 0.1268 ± 0.0060\clap{\,~$^\star$} & & 0.0841 ± 0.0010\clap{\,~$^\star$} & \\
&\hyperlink{cite.Keyulu18}{GINE}                         & 0.1265 ± 0.0076\clap{\,~$^\star$} & & 0.1339 ± 0.0044\clap{\,~$^\star$} &\\
&\hyperlink{cite.bresson2017residual}{GatedGCN}          & 0.2873 ± 0.0219\clap{\,~$^\star$} & & 0.2641 ± 0.0045\clap{\,~$^\star$} &\\
&\hyperlink{cite.dwivedi2020generalization}{Transformer} & 0.2694 ± 0.0098\clap{\,~$^\star$} & & 0.2618 ± 0.0031\clap{\,~$^\star$} &\\
&\hyperlink{cite.kreuzer2021rethinking}{SAN}             & 0.3230 ± 0.0039\clap{\,~$^\star$} & & 0.2592 ± 0.0158\clap{\,~$^\star$} &\\
&\hyperlink{cite.rampavsek2022recipe}{GPS}               & 0.3748 ± 0.0109\clap{\,~$^\star$} & & 0.3412 ± 0.0044\clap{\,~$^\star$} &\\
\midrule
\multirow{4}*{\rotatebox{90}{Ours}}&GCN                  & 0.2078 ± 0.0031           & \cellcolor{rwth-green!64}+64\% & 0.1338 ± 0.0007 & \cellcolor{rwth-green!59}+59\% \\
&GINE                                                    & 0.2718 ± 0.0054           & \cellcolor{rwth-green!100}+115\% & 0.2125 ± 0.0009 & \cellcolor{rwth-green!59}+59\% \\
&GatedGCN                                                & 0.3880 ± 0.0040           & \cellcolor{rwth-green!35}+35\% & 0.2922 ± 0.0018 & \cellcolor{rwth-green!11}+11\% \\
&GPS                                                     & 0.4440 ± 0.0065  & \cellcolor{rwth-green!18}+18\% & \textbf{0.3884 ± 0.0055} & \cellcolor{rwth-green!13}+13\% \\
\midrule
\multirow{3}*{\rotatebox{90}{Others}}&\hyperlink{cite.crawl}{\textsc{CRaWl}}     & \textbf{0.4588 ± 0.0079} & & - & \\
&\hyperlink{cite.pmlr-v202-gutteridge23a}{DRew}               & 0.3314 ± 0.0024\clap{\,~$^\star$} & & - & \\
&\hyperlink{cite.shirzad2023exphormer}{Exphormer}             & 0.3960 ± 0.0027\clap{\,~$^\star$} & & 0.3430 ±0.0008\clap{\,~$^\star$} & \\
\bottomrule
\end{tabular}
}
}
\end{small}
        \caption{Tuning results on vision datasets PascalVOC-SP and COCO-SP. $^\star$No normalization used.}
        \label{Tab:SP}
    \end{subtable}
    \hfill
    \begin{subfigure}{0.48\textwidth}
        \centering
        \includegraphics[width=\textwidth]{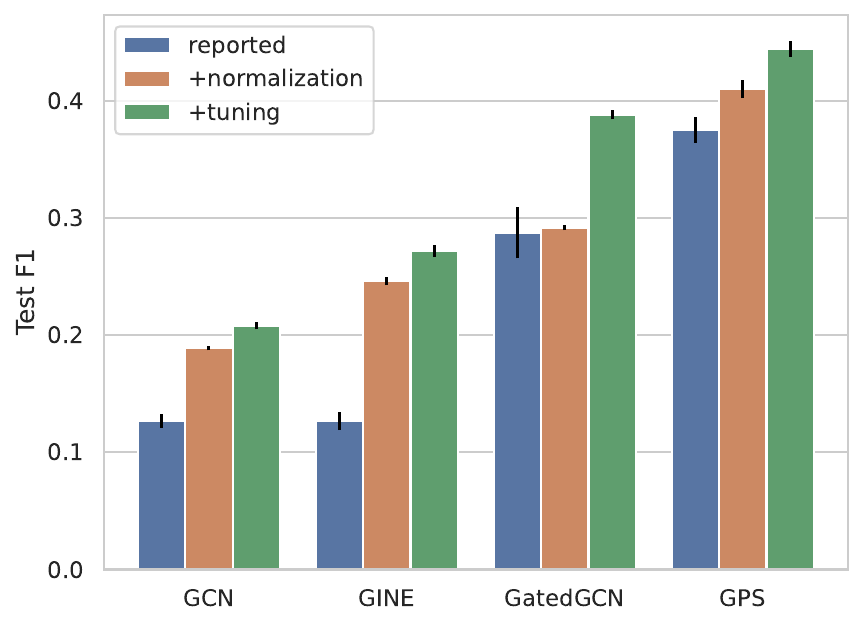}%
        \caption{The effects of feature normalization and hyperparameter tuning on PascalVOC-SP.}
        \label{Fig:PascalNorm}
    \end{subfigure}
    \caption{On PascalVOC-SP and COCO-SP feature normalization and further tuning improves performance across all compared methods.}
\end{figure}

\paragraph{Peptides-Func and Peptides-Struct}
Table \ref{peptides_results} provides the results obtained on the test splits of the Peptides-Func and Peptides-Struct.
For the MPGNN baselines we observe considerable improvements on both datasets as all three MPGNNs outperform GPS after tuning.
The average precision on Peptides-Func increased relatively by around 10\% to 20\%.
GCN achieves a score of 68.60\%, which is competitive with newer GTs such as GRIT or Graph ViT.
The improvement on Peptides-Struct is even more significant with a relative reduction of the MAE of 30\%, fully closing the gap to recently proposed GTs.
Surprisingly, a simple GCN is all you need to match the best known results on Peptides-Struct.
The results for GPS effectively stayed the same as in the original paper \cite{rampavsek2022recipe}. %
Those values thus seem to be representative for GPS.

We observed that the key hyperparameter underlying the improvements of all three MPGNNs is the depth of the prediction head.
To show this \cref{fig:peptides_head_ablation} contains an ablation where we exchanged the linear prediction head configured by \citet{dwivedi2022long} with a 2-layer perceptron, keeping all other hyperparameters the same.
While the benefit on Peptides-Func is considerable and highly significant, on Peptides-Struct the head depth accounts for almost the complete performance gap between MPGNNs and GTs.
GPS' performance with linear and deeper prediction heads is largely unchanged. 
For example, our GPS configurations in Table \ref{peptides_results} use a 2-layer prediction head. %
Our results indicate that the prediction targets of both datasets appear to depend non-linearly on global graph information.
In this case, MPGNNs with linear prediction heads are unable to model the target function.
Graph Transformers are not as sensitive to linear prediction heads, since each layer can process global graph information with a deep feed-forward network.
However, we would argue that switching to a deep predictive head represents a simpler and computationally cheaper solution to the same issue. %

\paragraph{PascalVOC-SP and COCO-SP}
Table \ref{Tab:SP} provides the results obtained on the test splits of the superpixel datasets PascalVOC-SP and COCO-SP.
We observe significant improvements for all evaluated methods.
On PascalVOC-SP the F1 score of GatedGCN increases to 38.80\% which exceeds the original performance reported for GPS by \citet{rampavsek2022recipe}.
GPS also improves significantly to 44.40\% F1.
This is only one percentage point below the results achieved by \textsc{CRaWl}, which currently is the only reported result with normalized features.
The previously large performance gap between GPS and \textsc{CRaWl} is therefore primarily explained by GPS processing raw input signals.
On COCO-SP, we observe similar results.
Here GPS sets a new state-of-the-art F1 score of 38.84\%.

Note that these improvements are achieved entirely through data normalization and hyperparameter tuning. 
\cref{Fig:PascalNorm} provides an ablation on the individual effect of normalization.
We train intermediate models with configurations identical to those used by \citet{dwivedi2022long} and \citet{rampavsek2022recipe}, but with feature normalization.
For GatedGCN we observe a slight performance increase but a large reduction in the variance across random seeds.  
For the remaining methods, including GPS, normalization of node and edge features already accounts for at least half of the observed performance gain, emphasizing its importance in practice. 

\paragraph{PCQM-Contact}
\setlength{\intextsep}{0pt plus 2pt minus 2pt}
\begin{wrapfigure}{T}{0.5\textwidth}
    \includegraphics[width=0.5\textwidth]{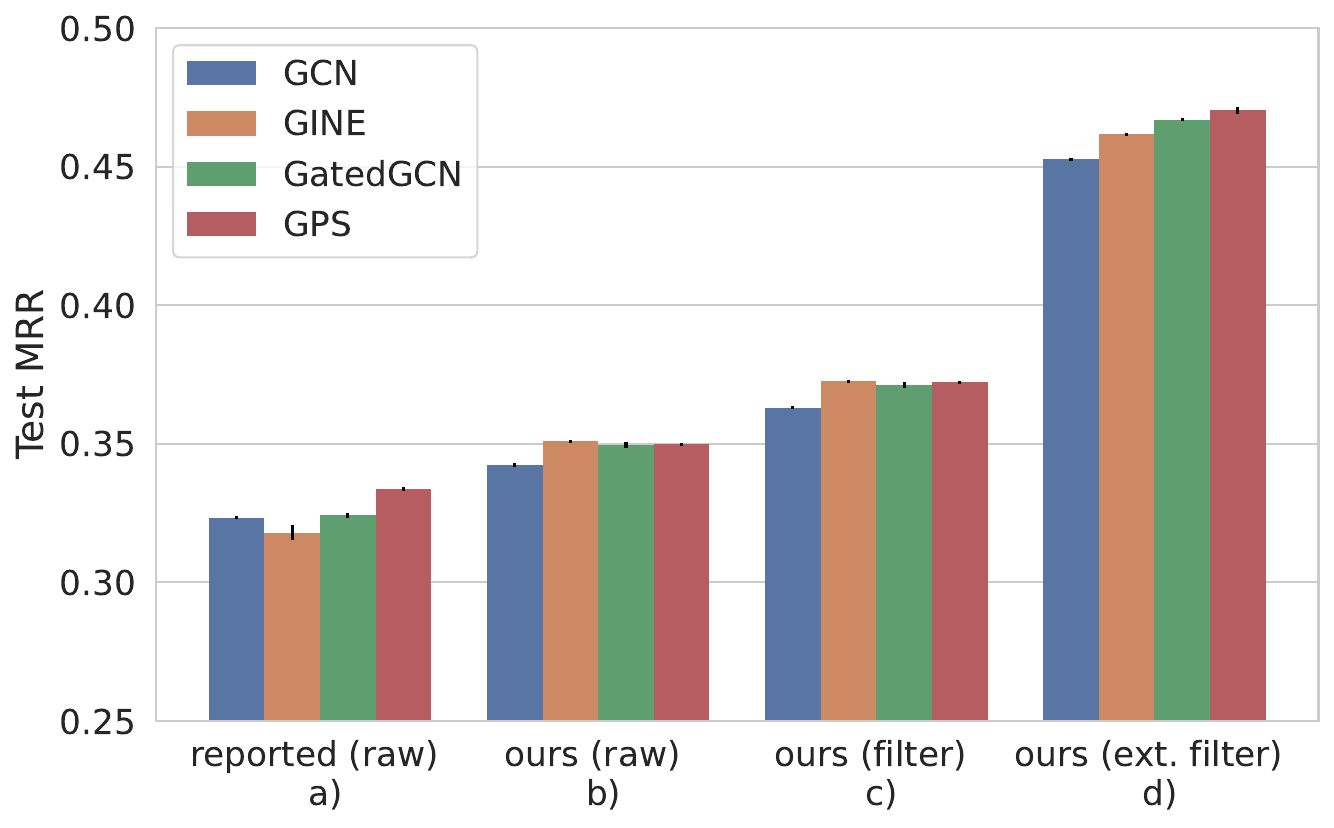}%
    \caption{Results on PCQM-Contact. For our own models we provide the MRR scores with varying levels of filtering.}
    \label{Fig:Contact}
\end{wrapfigure}
\cref{Fig:Contact} plots the MRR scores obtained on the test split with various evaluation settings as described in the link prediction paragraph of \cref{sec:concerns}.
First, we provide the results originally reported for LRGB in the literature (a). %
Recall that these values are obtained in a raw setting with false negatives present.
We then provide results obtained after training our own model with new hyperparameters (chosen based on the raw MRR) in b).
We still use the raw MRR for evaluation in b) to measure the impact of hyperparameter tuning.
Tuning yields an absolute improvement of around 3\%.
The previously reported slight performance edge of GPS is not observable in this setting after tuning. 

In subplot c) we measure the MRR of our models in the filtered setting.
Note that these values are based on the exact same predictions as in b), but false negatives are removed.
The measured MRR increases by roughly 3\% when compared to the raw setting.
This shift could erroneously be interpreted as a significant improvement when comparing to literature values obtained in a raw setting.
In d) we evaluate our models (still using the same predictions) in an extended filtered setting where we additionally remove self-loops from the candidate pool.
Compared to the filtered MRR in c) the MRR metric increases by about 10 percentage points, indicating that self-loops strongly affect the results.
Note that in d) GPS again slightly outperforms the MPGNN baselines, in contrast to b) and c).
This means that GPS' predictions seem to suffer overproportionally when self-loops are not filtered.
Therefore, the specific choice of how negative samples are filtered on PCQM-Contact can directly affect the ranking of compared methods and must be considered and implemented with care.

\section{Conclusion}
In our experiments we observed considerable performance gains for all three MPGNN baselines.
First, this indicates that extensive baseline tuning is important for properly assessing one's own method, escpecially on relatively recent datasets.
And second, only on the two superpixel datasets graph transformers exhibit clear performance benefits against MPGNNs, indicating that either there are ways to solve the other tasks without long-range interactions or graph transformers are not inherently better at exploiting such long-range dependencies.
Evaluating this further appears to be promising direction for future research. 
In addition, we would invite a discussion on the best-suited link prediction metric on PCQM-Contact.

\bibliographystyle{unsrtnat}
\bibliography{reference}

\appendix
\newpage

\section{Experiment Details}

\subsection{Hyperparameters}
\label{appendix:hp}

In the following we describe our methodology for tuning hyperparameters on the LRGB datasets.
We did not conduct a dense grid search, since this would be infeasible for all methods and datasets.
Instead we perform a ``linear'' hyperparameter search.
We start from a empricially chosen default config and tune each hyperparameter individually within a fixed range. %
Afterwards, we also evaluate the configuration obtained by combining the best choices of every hyperparameter.
From all tried configurations we then select the one with the best validation performance as our final setting.
For this hyperparameter sweep, we resorted to a single run per configuration and for the larger datasets slightly reduced the number of epochs. %
For the final evaluations runs we average results across four different random seeds as specified by the LRGB dataset.

Overall, we tried to incorporate the most important hyperparameters which we selected to be dropout, model depth, prediction head depth, learning rate, and the used positional or structural encoding.
For GPS we additionally evaluated the internal MPGNN (but only between GCN and GatedGCN) and whether to use BatchNorm or LayerNorm.
Thus, our hyperparamters and ranges were as follows:
\begin{itemize}
    \item Dropout [0, 0.1, 0.2], default 0.1
    \item Depth [6,8,10], default 8. The hidden dimension is chosen to stay within a hard limit of 500k parameters
    \item learning rate [0.001, 0.0005, 0.0001], default $0.001$
    \item Head depth [1,2,3], default 2
    \item Encoding [none, LapPE, RWSE] default none
    \item Internal MPGNN [GCN, GatedGCN], default GatedGCN (only for GPS)
    \item Normalization [BatchNorm, LayerNorm] default BatchNorm (only for GPS)
\end{itemize}
On the larger datasets PCQM-Contact and COCO we reduce the hyperparameters budget slightly for efficiency.
There, we did not tune the learning rate (it had been $0.001$ in every single other case) and omitted a dropout rate of 0.
We note that the tuning procedure used here is relatively simple and not exhaustive.
The ranges we searched are rather limited, especially in terms of network depth, and could be expanded in the future.
\cref{pepfunc_hp,pepstruct_hp,pascal_hp,coco_hp,pcqm_hp} provide all final model configurations after tuning. 
\cref{lrgb_results_full} provides the final performance on all datasets.

We make some additional setup changes based on preliminary experiments.
All models are trained with an AdamW optimizer using a cosine annealing learning rate schedule and linear warmup.
This differs from \citet{dwivedi2022long}, who optimized the MPGNN models with a ``Reduce on Plateau'' schedule and instead matches the learning rate schedule of GPS \cite{rampavsek2022recipe}. 
We set the weight decay to 0.0 in all five datasets and switch to slightly larger batch sizes to speed up convergence.
We also choose GeLU \cite{hendrycks2016gaussian} as our default activation function.
Furthermore, we change the prediction head for graph-level tasks such that all hidden layers have the same hidden dimension as the GNN itself.
These were previously configured to become more narrow with depth, but we could not observe any clear benefit from this design choice. 
Last, all MPGNN models use proper skip connections which go around the entire GNN layer.
The original LRGB results use an implementation of GCN as provided by GraphGym \cite{graphgym}.
The skip connections in this implementation do not skip the actual non-linearity at the end of each GCN layer, possibly hindering the flow of gradients.
We reimplement GCN with skip connections that go around the non-linearity.
Note that these additional tweaks are \textbf{not} used in our ablation studies in \cref{fig:peptides_head_ablation} and \cref{Fig:PascalNorm} when training the intermediate models where we only change the head depth and normalization, respectively.
There, we use identical model configurations to those used in the literature.

\subsection{Feature Normalization}
\label{appendix:norm}
On PascalVOC-SP and COCO-SP we apply channel-wise normalisation to the node and edge features.
For each dataset, we compute the channel-wise mean $\mu \in \mathbb{R}^d$ and standard deviation $\sigma \in \mathbb{R}^d$ on the train split.
Here, $d$ is the feature dimension.
Each feature vector $x \in \mathbb{R}^d$ is then normalized linearly before beigng passed to the model:
\begin{equation*}
    \Tilde{x}_i = \frac{x_i - \mu_i}{\sigma_i}
\end{equation*}

\newpage
\begin{table}[t]
\caption{
    Hyperparameters on Peptides-Func
}
\label{pepfunc_hp}
\begin{center}
\begin{small}
\resizebox{0.48\textwidth}{!}{
\begin{tabular}{lcccc}
\toprule
      & GCN & GINE & GatedGCN & GPS \\
\midrule
lr              & 0.001 & 0.001 & 0.001 & 0.001 \\
dropout         & 0.1 & 0.1 & 0.1 & 0.1 \\
\#layers        & 6 & 8 & 10 & 6 \\
hidden dim.     & 235 & 160 & 95 & 76 \\
head depth      & 3 & 3 & 3 & 2 \\
PE/SE           & RWSE & RWSE & RWSE & LapPE \\
batch size      & 200 & 200 & 200 & 200 \\
\#epochs        & 250 & 250 & 250 & 250 \\
norm            & - & - & - & BatchNorm \\
MPNN            & - & - & - & GatedGCN \\
\#Param.        & 486k & 491k & 493k & 479k \\
\end{tabular}
}
\end{small}
\end{center}
\vskip -0.1in
\end{table}

\begin{table}[t]
\caption{
    Hyperparameters on Peptides-Struct.
}
\label{pepstruct_hp}
\begin{center}
\begin{small}
\resizebox{0.48\textwidth}{!}{
\begin{tabular}{lcccc}
\toprule
      & GCN & GINE & GatedGCN & GPS \\
\midrule
lr              & 0.001 & 0.001 & 0.001 & 0.001 \\
dropout         & 0.1 & 0.1 & 0.1 & 0.1 \\
\#layers        & 6 & 10 & 8 & 8 \\
hidden dim.     & 235 & 145 & 100 & 64 \\
head depth      & 3 & 3 & 3 & 2 \\
PE/SE           & LapPE & LapPE & LapPE & LapPE \\
batch size      & 200 & 200 & 200 & 200 \\
\#epochs        & 250 & 250 & 250 & 250 \\
norm            & - & - & - & BatchNorm \\
MPNN            & - & - & - & GatedGCN \\
\#Param.        & 488k & 492k & 445k & 452k \\
\end{tabular}
}
\end{small}
\end{center}
\vskip -0.1in
\end{table}

\begin{table}[t]
\caption{
    Hyperparameters on PascalVOC-SP.
}
\label{pascal_hp}
\begin{center}
\begin{small}
\resizebox{0.48\textwidth}{!}{
\begin{tabular}{lcccc}
\toprule
      & GCN & GINE & GatedGCN & GPS \\
\midrule
lr              & 0.001 & 0.001 & 0.001 & 0.001 \\
dropout         & 0.0 & 0.2 & 0.2 & 0.1 \\
\#layers        & 10 & 10 & 10 & 8 \\
hidden dim.     & 200 & 145 & 95 & 68 \\
head depth      & 3 & 2 & 2 & 2 \\
PE/SE           & RWSE & none & none & LapPE \\
batch size      & 50 & 50 & 50 & 50 \\
\#epochs        & 200 & 200 & 200 & 200 \\
norm            & - & - & - & BatchNorm \\
MPNN            & - & - & - & GatedGCN \\
\#Param.        & 490k & 450k & 473k & 501k \\
\end{tabular}
}
\end{small}
\end{center}
\vskip -0.1in
\end{table}

\begin{table}[t]
\caption{
    Hyperparameters on COCO-SP.
}
\label{coco_hp}
\begin{center}
\begin{small}
\resizebox{0.48\textwidth}{!}{
\begin{tabular}{lcccc}
\toprule
      & GCN & GINE & GatedGCN & GPS \\
\midrule
lr              & 0.001 & 0.001 & 0.001 & 0.001 \\
dropout         & 0.1 & 0.1 & 0.1 & 0.1 \\
\#layers        & 6 & 6 & 8 & 8 \\
hidden dim.     & 280 & 195 & 105 & 68 \\
head depth      & 1 & 1 & 1 & 1 \\
PE/SE           & none & none & none & none \\
batch size      & 50 & 50 & 50 & 50 \\
\#epochs        & 200 & 200 & 200 & 200 \\
norm            & - & - & - & LayerNorm \\
MPNN            & - & - & - & GatedGCN \\
\#Param.        & 500k & 478k & 459k & 500k \\
\end{tabular}
}
\end{small}
\end{center}
\vskip -0.1in
\end{table}

\begin{table}[t]
\caption{
    Hyperparameters on PCQM-Contact.
}
\label{pcqm_hp}
\begin{center}
\begin{small}
\resizebox{0.48\textwidth}{!}{
\begin{tabular}{lcccc}
\toprule
      & GCN & GINE & GatedGCN & GPS \\
\midrule
lr              & 0.001 & 0.001 & 0.001 & 0.001 \\
dropout         & 0.1 & 0.1 & 0.1 & 0.0 \\
\#layers        & 8 & 8 & 8 & 6 \\
hidden dim.     & 215 & 160 & 105 & 76 \\
head depth      & 1 & 1 & 1 & 1 \\
PE/SE           & LapPE & LapPE & LapPE & LapPE \\
batch size      & 500 & 500 & 500 & 500 \\
\#epochs        & 150 & 150 & 150 & 150 \\
norm            & - & - & - & LayerNorm \\
MPNN            & - & - & - & GatedGCN \\
\#Param.        & 456k & 466k & 477k & 478k \\
\end{tabular}
}
\end{small}
\end{center}
\vskip -0.1in
\end{table}

\begin{table*}[t]
\caption{
    Performance of our models on the Long-Range Graph Benchmark.
}
\label{lrgb_results_full}
\begin{center}
\begin{small}
\resizebox{\textwidth}{!}{
\textsc{
\begin{tabular}{lccccccc}
\toprule
Method      & PASCALVOC-SP & COCO-SP & PEPTIDES-FUNC & PEPTIDES-STRUCT & \multicolumn{3}{c}{PCQM-CONTACT} \\
            & Test F1 $\uparrow$ & Test F1 $\uparrow$ & Test AP $\uparrow$ & Test MAE $\downarrow$ & \multicolumn{3}{c}{Test MRR $\uparrow$} \\
            & & & & & raw & filter & ext. filter \\
\midrule
GCN         & 0.2078 ± 0.0031 & 0.1338 ± 0.0007 & 0.6860 ± 0.0050 & 0.2460 ± 0.0007 & 0.3424 ± 0.0007 & 0.3631 ± 0.0006 & 0.4526 ± 0.0006 \\
GINE        & 0.2718 ± 0.0054 & 0.2125 ± 0.0009 & 0.6621 ± 0.0067 & 0.2473 ± 0.0017 & 0.3509 ± 0.0006 & 0.3725 ± 0.0006 & 0.4617 ± 0.0005 \\
GatedGCN    & 0.3880 ± 0.0040 & 0.2922 ± 0.0018 & 0.6765 ± 0.0047 & 0.2477 ± 0.0009 & 0.3495 ± 0.0010 & 0.3714 ± 0.0010 & 0.4670 ± 0.0004 \\
GPS         & 0.4440 ± 0.0065 & 0.3884 ± 0.0055 & 0.6534 ± 0.0091 & 0.2509 ± 0.0014 & 0.3498 ± 0.0005 & 0.3722 ± 0.0005 & 0.4703 ± 0.0014 \\
\bottomrule
\end{tabular}
}
}
\end{small}
\end{center}
\vskip -0.1in
\end{table*}

\end{document}